\documentclass[journal]{IEEEtran}
\ifCLASSINFOpdf
\else
\fi

\usepackage[dvipsnames]{xcolor}
\usepackage{amsmath,amssymb,amsfonts}
\usepackage{graphicx}
\usepackage{url}
\usepackage{xcolor}
\usepackage{multirow} % for multirow feature
\usepackage{multicol} % for multicolumn feature
\usepackage{makecell} % Required for \Xhline
\usepackage{enumitem}
\usepackage{colortbl}
\usepackage{nicematrix}
\usepackage{marvosym}
\begin{document}
%
% paper title
% Titles are generally capitalized except for words such as a, an, and, as,
% at, but, by, for, in, nor, of, on, or, the, to and up, which are usually
% not capitalized unless they are the first or last word of the title.
% Linebreaks \\ can be used within to get better formatting as desired.
% Do not put math or special symbols in the title.
\title{LYT-NET: Lightweight YUV Transformer-based Network for Low-light Image Enhancement}
%
%
% author names and IEEE memberships
% note positions of commas and nonbreaking spaces ( ~ ) LaTeX will not break
% a structure at a ~ so this keeps an author's name from being broken across
% two lines.
% use \thanks{} to gain access to the first footnote area
% a separate \thanks must be used for each paragraph as LaTeX2e's \thanks
% was not built to handle multiple paragraphs
%

\author{\IEEEauthorblockN{Alexandru Brateanu\IEEEauthorrefmark{1}, Raul Balmez\IEEEauthorrefmark{1}, Adrian Avram\IEEEauthorrefmark{2}, Ciprian Orhei\IEEEauthorrefmark{2}\IEEEauthorrefmark{3} and Cosmin Ancuti\IEEEauthorrefmark{2}\IEEEauthorrefmark{3}}\\
\IEEEauthorblockA{\IEEEauthorrefmark{1}University of Manchester, United Kingdom\\}
\IEEEauthorblockA{\IEEEauthorrefmark{2}University Politehnica Timisoara, Romania\\}
\IEEEauthorblockA{\IEEEauthorrefmark{3}West University of Timisoara, Romania\\}
}
% <-this % stops a space
%\thanks{J. Doe and J. Doe are with Anonymous University.}% <-this % stops a space
%\thanks{Manuscript received September, 2024; revised xxx, 2024.}}

% note the % following the last \IEEEmembership and also \thanks - 
% these prevent an unwanted space from occurring between the last author name
% and the end of the author line. i.e., if you had this:
% 
% \author{....lastname \thanks{...} \thanks{...} }
%                     ^------------^------------^----Do not want these spaces!
%
% a space would be appended to the last name and could cause every name on that
% line to be shifted left slightly. This is one of those "LaTeX things". For
% instance, "\textbf{A} \textbf{B}" will typeset as "A B" not "AB". To get
% "AB" then you have to do: "\textbf{A}\textbf{B}"
% \thanks is no different in this regard, so shield the last } of each \thanks
% that ends a line with a % and do not let a space in before the next \thanks.
% Spaces after \IEEEmembership other than the last one are OK (and needed) as
% you are supposed to have spaces between the names. For what it is worth,
% this is a minor point as most people would not even notice if the said evil
% space somehow managed to creep in.

% The paper headers
\markboth{LYT-NET: Lightweight YUV Transformer-based Network for Low-light Image Enhancement}%
{Brateanu \MakeLowercase{\textit{et al.}}: LYT-NET: Lightweight YUV Transformer-based Network for Low-light Image Enhancement}
% The only time the second header will appear is for the odd numbered pages
% after the title page when using the twoside option.
% 
% *** Note that you probably will NOT want to include the author's ***
% *** name in the headers of peer review papers.                   ***
% You can use \ifCLASSOPTIONpeerreview for conditional compilation here if
% you desire.

% If you want to put a publisher's ID mark on the page you can do it like
% this:
%\IEEEpubid{0000--0000/00\$00.00~\copyright~2015 IEEE}
% Remember, if you use this you must call \IEEEpubidadjcol in the second
% column for its text to clear the IEEEpubid mark.

% use for special paper notices
%\IEEEspecialpapernotice{(Invited Paper)}

% make the title area
\maketitle

\begin{abstract}
%zoinks
This letter introduces LYT-Net, a novel lightweight transformer-based model for low-light image enhancement. LYT-Net consists of several layers and detachable blocks, including our novel blocks—Channel-Wise Denoiser (\textbf{CWD}) and Multi-Stage Squeeze \& Excite Fusion (\textbf{MSEF})—along with the traditional Transformer block, Multi-Headed Self-Attention (\textbf{MHSA}). In our method we adopt a dual-path approach, treating chrominance channels \( U \) and \( V \) and luminance channel \( Y \)   as separate entities to help the model better handle illumination adjustment and corruption restoration.  Our comprehensive evaluation on established LLIE datasets demonstrates that, despite its low complexity, our model outperforms recent LLIE methods.
The source code and pre-trained models are available at \color{purple}{\url{https://github.com/albrateanu/LYT-Net}}
\end{abstract}

% Note that keywords are not normally used for peerreview papers.

\section{Introduction}
Low-light image enhancement (LLIE) is an important and challenging task in computational imaging. When images are captured in low-light conditions, their quality often deteriorates, leading to a loss of detail and contrast. This not only makes the images visually unappealing but also affects the performance of many imaging systems. The goal of LLIE is to improve the clarity and contrast of these images, while also correcting distortions that commonly occur in dark environments, all without introducing unwanted artifacts or causing imbalances in color.

\vspace{-0.5mm}
%zoinks

         Earlier LLIE methods~\cite{HE_review} primarily relied on frequency decomposition~\cite{Xiao_2016, Kim_2016}, histogram equalization~\cite{ Chen_2003, Kansal_2018}, and Retinex theory~\cite{Retinex, Park_2017, Gu_2019, Jang_2012}. With the rapid advancement of deep learning, various CNN architectures~\cite{RetinexNet, DeepUPE, Diff-retinex, KinD, Zhang_2020, Dudhane_2022, MIRNet,RUAS,SNR-Net, shi2024zero} have been shown to outperform traditional LLIE techniques. Based on Retinex theory, Retinex-Net~\cite{RetinexNet} integrates Retinex decomposition with an original CNN architecture, while Diff-Retinex~\cite{Diff-retinex} proposes a generative framework to further address content loss and color deviation caused by low light.

 \begin{figure}[h!]
        \centering
        \setlength{\tabcolsep}{0.5pt}

        \includegraphics[width=\linewidth]{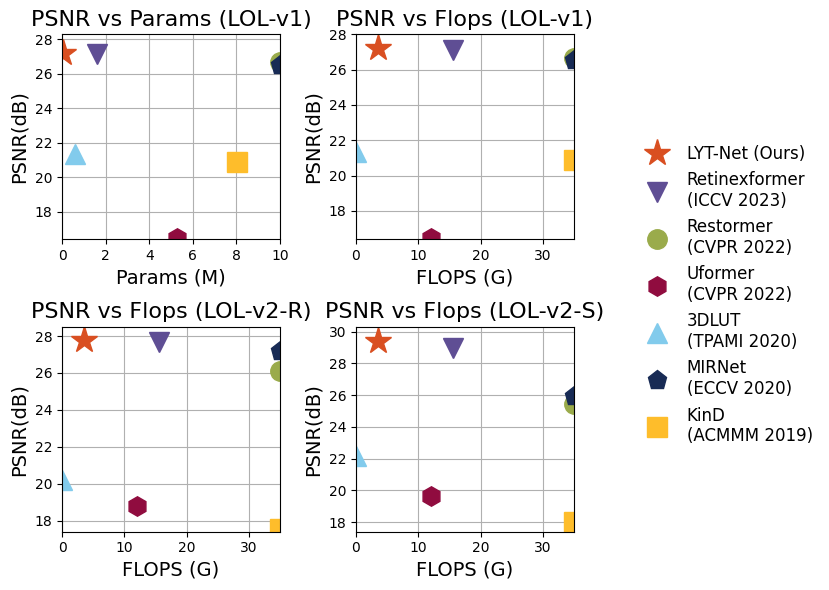}
        
       \caption{Our model delivers SOTA performance in LLIE task, while maintaining computational efficiency (results are plotted on LOL dataset \cite{RetinexNet}).
       }
        \label{fig:SOTA_overview}
        \vspace{-3.4mm}
    \end{figure}
\vspace{-0.5mm}
%zoinks

    The development of Generative Adversarial Networks (GAN) \cite{Goodfellow_2014} has provided a new perspective for LLIE, where low-light images are used as input to generate their normal-light counterparts. For instance, EnlightenGAN \cite{EnGAN} employs a single generator model to directly convert low-light images to normal-light versions, effectively using both global and local discriminators in the transformation process. %SNR-Net \cite{SNR-Net} explores the relationship between signal and noise in the image space, utilizing the SNR for spatially varying enhancement.

\vspace{-0.5mm}
%zoinks

More recently, Vision Transformers (ViTs)~\cite{Dosovitskiy_2021} have demonstrated significant effectiveness in various computer vision tasks~\cite{wang2023ultra, Wang_ICCV_2021, Zheng_CVPR_2021, Liu_ICCV_2021}, largely due to the self-attention (SA) mechanism. Despite these advancements, the application of ViTs to low-level vision tasks remains relatively underexplored. Only a few LLIE-ViT-based strategies have been introduced in the recent literature~\cite{UFormer, Retinexformer, Restormer, jte-cflow}.
% Uformer~\cite{UFormer} is based on the classical UNet architecture, where the convolution layers are replaced with Transformer blocks while maintaining the hierarchical encoder-decoder structure and skip connections. 
Restormer~\cite{Restormer}, on the other hand, introduces a multi-Dconv head transposed attention (MDTA) block, replacing the vanilla multi-head self-attention.

    \begin{figure*}[ht!]
        \centering
        \includegraphics[width=1.0\linewidth]{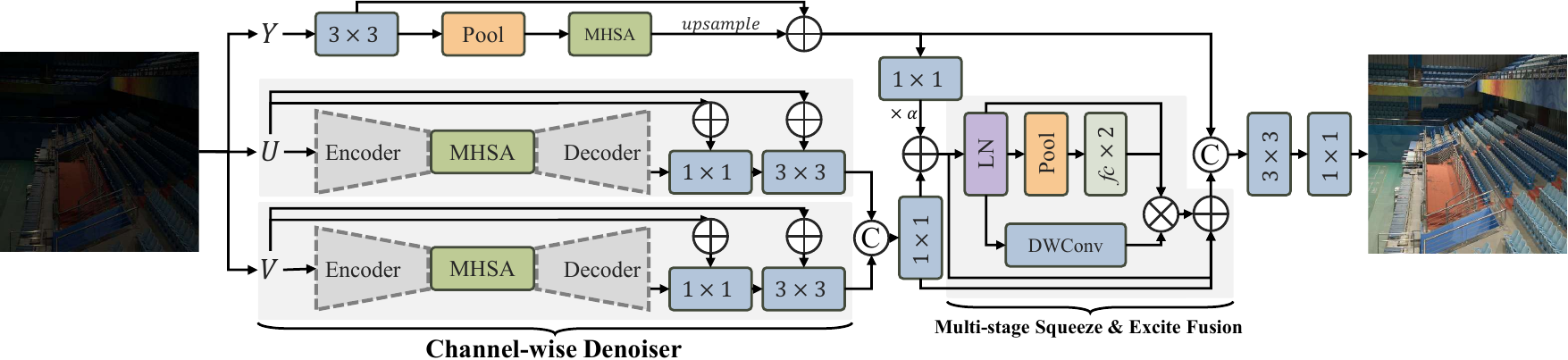}

        \caption{ Overall framework of \textbf{LYT-Net}. The architecture consists of several detachable blocks like Channel-wise Denoiser (CWD), Multi-headed Self-Attention (MHSA), Multi-stage Squeeze and Excite Fusion (MSEF). }
        \label{fig:proposed_method}
        \vspace{-5mm}
    \end{figure*}

% \textcolor{blue}{
Diffusion models have emerged as a powerful approach for LLIE, leveraging their ability to learn complex data distributions through a simulated forward process \cite{jiang2023low, jiang2025lightendiffusion, hou2024global}. 
% }

    %In this paper, we propose a novel transformer-based approach, characterized by its low-weight design, that achieves state-of-the-art (SOTA) results in LLIE while maintaining computational efficiency. In Fig. \ref{fig:SOTA_overview} we present a comparative analysis of performance over complexity between SOTA methods evaluated using the LOL dataset \cite{RetinexNet}. For visual clarity, in the short comparison, we omitted models such as Restormer\cite{Restormer} and MIRNet\cite{MIRNet} as they utilize considerably more parameters yet yield inferior results.
\vspace{-0.5mm}
%blue

In this letter, we propose a novel lightweight transformer-based approach called \textbf{LYT-Net}. Different from the existing transformer-based methods, our method focuses on computational efficiency while still producing state-of-the-art (SOTA) results. Specifically,  we first separate chrominance from luminance employing the YUV color space. The chrominance information (channels \( U \) and \( V \)) is initially processed through a specialized Channel-wise Denoiser (\textbf{CWD}) block, which reduces noise while preserving fine details. To minimize computational complexity, the luminance channel \( Y \) undergoes convolution and pooling to extract features, which are subsequently enhanced by a traditional Multi-headed Self-Attention (\textbf{MHSA}) block. The enhanced channels are then recombined and processed through a novel Multi-stage Squeeze and Excite Fusion (\textbf{MSEF}) block. Finally, the chrominance channels \( U \) and \( V \) channels are concatenated with the luminance \( Y \)  channel and passed through a final set of convolutional layers to produce the restored image.

\vspace{-0.5mm}
%zoinks

Our method has undergone extensive testing on established LLIE datasets. Both qualitative and quantitative evaluations indicate that our approach achieves highly competitive results.  Fig. \ref{fig:SOTA_overview} presents  a comparative analysis of performance over complexity between SOTA methods evaluated using the LOL dataset \cite{RetinexNet}. It can be observed that, despite its lightweight design, our method produces results that are not only comparable to, but often outperform, those of more complex recent deep learning LLIE techniques.
\vspace{-3mm}

    %\textbf{LYT-Net}, demonstrates encouraging results, as illustrated in Fig. \ref{fig:visual_hist}. Here, we have plotted the histogram of the predicted image to provide a clearer understanding of the model's effect. Notably, the distribution in the histogram is smoother compared to the ground truth, which is a favorable outcome. This smoother distribution indicates that \textbf{LYT-Net} effectively enhances the image, balancing its tonal range and improving overall visual quality.

    %Our proposed model utilizes the YUV color space, which is particularly advantageous for LLIE due to its distinct separation of luminance (Y) and chrominance (U and V). By employing this color space, we can specifically target enhancements that can improve image visibility and detail in low-light conditions without adversely affecting the color information. Since human vision is more sensitive to changes in luminance, focusing on the Y channel leads to more natural and perceptually appealing enhancements.

    %The paper is structured as follows: Section \ref{sec:intro} provides an overview of the research and its importance. Section \ref{sec:relate}, "Related Work," aims to provide an overview on of the LLIE domain. In Section \ref{sec:method}, "Proposed Method," we detail the proposed model and its key aspects, while in Section \ref{sec:experiments} we present the results on the LOL datasets. The paper concludes with Section \ref{sec:conclusion}, summarizing the findings and suggesting future research directions.

%\section{Related Work}
%\input{related.tex}
\section{Our Approach}
% \vspace{-1mm}/

%\section{PROPOSED METHOD}
\label{sec:method}

% \textcolor{red}{TOTAL max 3 coloane}

    \begin{figure*}[!h]
            \centering
            \includegraphics[width=\linewidth]{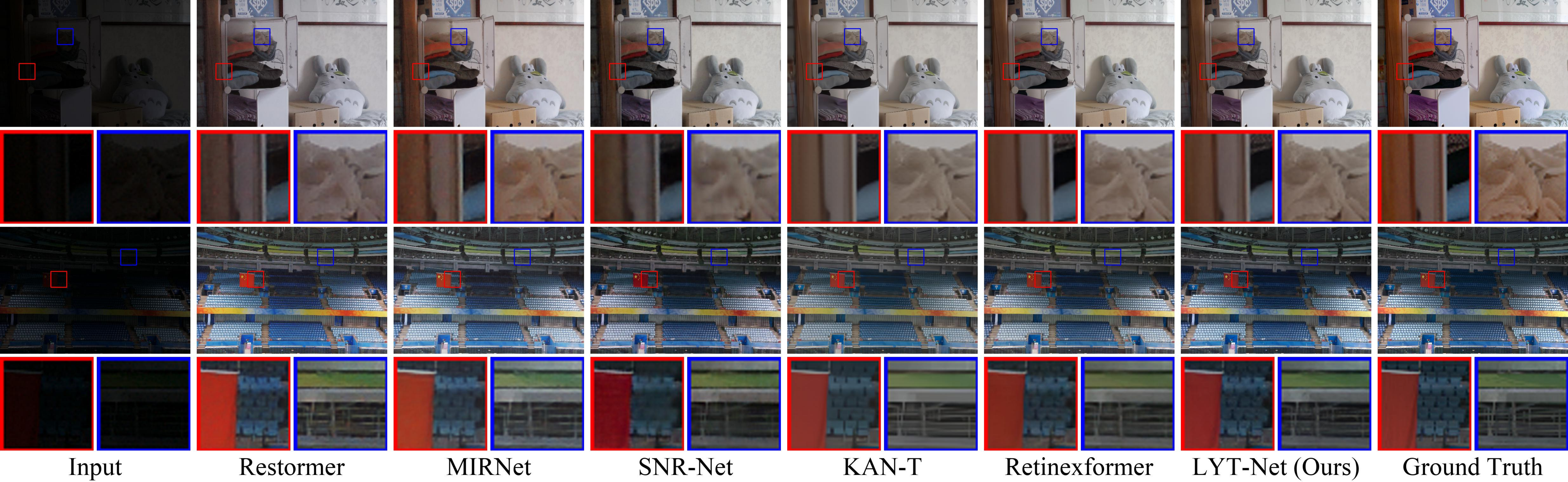}            
            \centering
            \vspace{-20pt}
            \caption{\centering Qualitative comparison with SOTA LLIE methods on the LOL dataset. Zoom-in regions are used to illustrate differences.}
            \label{fig:compare_LOL_v2_zoom}
            \vspace{-5mm}
    \end{figure*}

    %In Fig. \ref{fig:proposed_method} we illustrate the overall architecture of \textbf{LYT-Net}. As shown, the model consists of a main YUV decomposition to separate chrominance from luminance, followed by several layers and detachable blocks, like the Multi-headed Self-attention (MHSA) Block, Multi-stage Squeeze \& Excite Fusion (MSEF) Block and Channel-wise Denoiser (CWD) Block. We adopt a dual-path approach, focusing on chrominance and luminance as separate entities, to help the model better understand the between difference illumination adjustment and corruption restoration.

    %As seen in Fig. \ref{fig:proposed_method}, the model processes an input image in RGB format and converts it into YUV. Each channel is individually enhanced using a series of convolutional layers, pooling operations, and the MHSA mechanism. The luminance channel \( Y \) undergoes convolution and pooling to extract features, which are then enhanced by the MHSA block. Chrominance channels \( U \) and \( V \) are processed through a CWD block to reduce noise while preserving details. The enhanced  chrominance channels are then recombined and processed through the MSEF block. Ultimately, chrominance \(UV\) and luminance \( Y \) are concatenated and fed through a final set of convolutional layers to produce the output, yielding a high-quality enhanced image.
 
%zoinks

In Fig. \ref{fig:proposed_method}, we illustrate the overall architecture of \textbf{LYT-Net}, which consists of several layers and detachable blocks, including our novel blocks—Channel-Wise Denoiser (\textbf{CWD}) and Multi-Stage Squeeze \& Excite Fusion (\textbf{MSEF})—along with the traditional ViT block, Multi-Headed Self-Attention (\textbf{MHSA}). We adopt a dual-path approach, treating chrominance and luminance as separate entities to help the model better handle illumination adjustment and corruption restoration. The luminance channel \( Y \) undergoes convolution and pooling to extract features, which are then enhanced by the \textbf{MHSA} block. Chrominance channels \( U \) and \( V \) are processed through the \textbf{CWD} block to reduce noise while preserving details. The enhanced chrominance channels are then recombined and processed through the \textbf{MSEF} block. Finally, the chrominance \(U,V\) and luminance \( Y \)  channels are concatenated and passed through a final set of convolutional layers to produce the output, resulting in a high-quality, enhanced image.

    % \subsection{Multi-headed Self-attention Block}
    % Previous transformer implementations propose the transformer block as consisting of two Layer Normalization (LN) layers, MHSA block and a Feed-forward Network (FFN). Due to the inherent computational complexity of transformer blocks, we adopt a simplified version, utilizing the MHSA exclusively.
    % Firstly, the input feature \( F_{\text{in}} \in \mathbb{R}^{H \times W \times C} \) is projected linearly into query elements \( Q \), key elements \( K \), and value elements \( V \) through bias-free, fully connected layers. This projection is mathematically represented as: \[ Q = X W_{Q}^{T}, K = X W_{K}^{T}, V = X W_{V}^{T}, Q, K, V \in \mathbb{R}^{HW \times C} \] where \( W_{Q} \), \( W_{K} \) and \( W_{V} \) denote learnable parameters of the dense layers, and \( T \) denotes the matrix transpose. 
    % Then, \( Q \), \( K \), and \( V \) are reshaped into a series of tokens \( X \), which are then are divided into \( k \) heads formulated as: \[ X = [X_{1}, \ldots, X_{k}], X_{i} \in \mathbb{R}^{HW \times d_{k}}, d_{k} = \frac{C}{k}, i=\bar{1,k} \]
    % The self-attention for each head \( i \) is then defined as: \[\text{Attention}(Q_{i}, K_{i}, V_{i}) = V_{i} \odot \text{Softmax}(K_{i}^{T} Q_{i})\] The outputs of the \( k \) heads are concatenated and fed through a fully-connected layer, added to a positional encoding \( P \) with learnable parameters to yield the output tokens \( X_{\text{out}} \), which is reshaped to produce the output feature \( F_{\text{out}} \in  \mathbb{R}^{H \times W \times C}\).

 \subsection{Channel-wise Denoiser Block}

        % The CWD integrates convolutional and attention-based mechanisms for image denoising. It consists of a U-shaped network that employs a MHSA Block as the bottleneck. This allows for efficient feature-capture at different scales while preserving detail information through the use of skip connections.
        % It consists of a series of four conv3x3. The first conv3x3 has strides=1 for feature extraction. The other three conv3x3 layers have strides=2, helping with capturing features at different scales. Then, the integration of the attention bottleneck enables the model to capture long-range dependencies, which is essential in tasks like denoising where context plays a key role. Then the image is reconstructed through the use of upsampling layers and skip connections to facilitate recovery of spatial resolution and detail. Lastly, the final layer reconstructs the denoised image while also incorporating a residual connection to aid in preserving low-level details of the image, preventing the denoiser from over-smoothing out important features.

        \vspace{0.5mm}
        The \textbf{CWD} Block employs a U-shaped network with \textbf{MHSA} as the bottleneck, integrating convolutional and attention-based mechanisms. It includes multiple \textit{conv3$\times$3} layers with varying strides and skip connections, facilitating detailed feature capture and denoising.
        
        \vspace{-0.5mm}
        It consists of a series of four \textit{conv3$\times$3} layers. The first \textit{conv3$\times$3} has strides of $1$ for feature extraction. The other three \textit{conv3$\times$3} layers have strides of $2$, helping with capturing features at different scales. The integration of the attention bottleneck enables the model to capture long-range dependencies, followed by upsampling layers and skip connections to reconstruct and facilitate the recovery of spatial resolution.

        \vspace{-0.5mm}
        %eu
        \textcolor{black}{
        This approach allows us to apply \textbf{MHSA} on a feature map with reduced spatial dimensions, significantly improving computational efficiency. Additionally, using interpolation-based upsampling instead of transposed convolutions cuts the number of parameters in the CWD by more than half, while preserving performance.}

    \subsection{Multi-headed Self-attention Block}

        In our updated simplified transformer architecture, the input feature $\mathbf{F}_{\text{in}} \in \mathbb{R}^{H \times W \times C}$ is first linearly projected into query ($\mathbf{Q}$), key ($\mathbf{K}$), and value ($\mathbf{V}$) components through bias-free fully connected layers. The linear layers use parameter $D$ to determine projection head dimensionality.
        
        \vspace{-0.5mm}
        \begin{equation}
        \small
        \mathbf{Q} = \mathbf{X} \mathbf{W}_{\mathbf{Q}}, \mathbf{K} = \mathbf{X} \mathbf{W}_{\mathbf{K}}, \mathbf{V} = \mathbf{X} \mathbf{W}_{\mathbf{V}},~ \mathbf{Q}, \mathbf{K}, \mathbf{V} \in \mathbb{R}^{HW \times D}
        \label{linear_proj}
        \end{equation}
        % \vspace{0.5mm}

        where $\mathbf{W}_{\mathbf{Q}}, \mathbf{W}_{\mathbf{K}}, \mathbf{W}_{\mathbf{V}}$ are fully connected layer weights. Next, these projected features are split into $k$ heads as such:
        \vspace{-0.5mm}
        \begin{equation}
        \small
        \mathbf{X} = [\mathbf{X}_1,\mathbf{X}_2,\cdots,\mathbf{X}_k],~ \mathbf{X}_i \in \mathbb{R}^{HW \times d_k},~ d_k = \frac{D}{k}, i = \overline{1, k}
        \label{split_heads}
        \vspace{-0.5mm}
        \end{equation}
        
        where each head operates independently with dimensionality $d_k$. The self-attention mechanism is applied to each head, as defined below:
        \vspace{-0.5mm}
        \begin{equation}
        \small
        \text{Attention}(\mathbf{Q}_i, \mathbf{K}_i, \mathbf{V}_i) = \text{Softmax}\left(\frac{\mathbf{Q}_i \mathbf{K}_i^\text{T}}{\sqrt{d_k}}\right)\!\times\!\mathbf{V}_i
        \label{attention}
        \vspace{-0.5mm}
        \end{equation}
        
        Finally, the attention outputs from all heads are concatenated and the combined output is passed through a linear layer to project it back to the original embedding size. The output tokens $\mathbf{X}_{\text{out}}$ are reshaped back into the original spatial dimensions to form the output feature $\mathbf{F}_{\text{out}} \in \mathbb{R}^{H \times W \times C}$.

        % \subsection{Multi-stage Squeeze \& Excite Fusion Block}
        % The Multi-stage Squeeze \& Excite Fusion (MSEF) Block processes an input feature \( F_{\text{in}} \in \mathbb{R}^{H \times W \times C} \) to enhance spatial and channel-wise features. Initially, \( F_{\text{in}} \) undergoes layer normalization, followed by global average pooling to capture global spatial context, producing a reduced descriptor \( S_{\text{reduced}} \). This descriptor is then expanded back to the original dimensions through fully-connected layers with ReLU and Tanh activations, resulting in \( S_{\text{expanded}} \).
        
        % % \[
        % % S_{\text{reduced}} = \text{ReLU}(W_1 \cdot \text{GlobalAvgPool}(\text{LayerNorm}(F_{\text{in}})))
        % % \]
        % % \[
        % % S_{\text{expanded}} = \text{Tanh}(W_2 \cdot S_{\text{reduced}})
        % % \]
        
        % The expanded feature \( S_{\text{expanded}} \) is reshaped and undergoes a point-wise multiplication with the output of a depthwise convolution layer applied to \( F_{\text{in}} \), denoted as \( Y_{\text{dwc}} \). The fused output \( X_{\text{fused}} \) is obtained by combining the scaled and spatially transformed features:
        
        % % \[
        % % X_{\text{fused}} = (\text{Reshape}(S_{\text{expanded}}) \odot Y_{\text{dwc}})
        % % \]
        
        % A residual connection is added by summing \( X_{\text{fused}} \) with the original input \( F_{\text{in}} \), yielding the final output feature map \( F_{out}\). 
        
        % The MSEF Block effectively integrates channel-wise recalibrated features with spatially enhanced features while preserving original input characteristics through the residual connection.

    \subsection{Multi-stage Squeeze \& Excite Fusion Block}

        \vspace{0.5mm}
        The \textbf{MSEF} Block enhances both spatial and channel-wise features of $\mathbf{F}_{\text{in}}$. Initially, $\mathbf{F}_{\text{in}}$ undergoes layer normalization, followed by global average pooling to capture global spatial context and a reduced fully-connected layer with ReLU activation, producing a reduced descriptor $\mathbf{S}_{\text{reduced}}$, as shown in Eq.~\eqref{formula:msef_reduce}. This descriptor is then expanded back to the original dimensions through another fully-connected layer with Tanh activation, resulting in $\mathbf{S}_{\text{expanded}}$, Eq.~\eqref{formula:msef_expand}.

        \vspace{-0.5mm}
        %eu
        \textcolor{black}{
        These operations compress the feature map into a reduced descriptor (the \textbf{squeezing} operation) to capture essential details, and then re-expand it (the \textbf{excitation} operation) to restore the full dimensions, emphasizing the most relevant features.}
        
        \vspace{-10pt}
        \begin{equation}
        \small
        \mathbf{S}_{\text{reduced}} = \text{ReLU}(\mathbf{W}_1 \cdot \text{GlobalPool}(\text{LayerNorm}(\mathbf{F}_{\text{in}})))
        \label{formula:msef_reduce}
        \end{equation}
        \vspace{-11pt}

        \vspace{-11pt}
        \begin{equation}
        \small
        \mathbf{S}_{\text{expanded}} = \text{Tanh}(\mathbf{W}_2 \cdot \mathbf{S}_{\text{reduced}}) \cdot \text{LayerNorm}(\mathbf{F}_{\text{in}})
        \label{formula:msef_expand}
        \end{equation}
        \vspace{-10pt}
        
        A residual connection is added to the fused output to produce the final output feature map $\mathbf{F}_{\text{out}}$, as in Eq.~\eqref{formula:msef_final}.
        \begin{equation}
        \small
        \mathbf{F}_{\text{out}} = \text{DWConv}(\text{LayerNorm}(\mathbf{F}_{\text{in}})) \cdot \mathbf{S}_{\text{expanded}} + \mathbf{F}_{\text{in}}
        \label{formula:msef_final}
        \end{equation}
        \vspace{-5pt}

        % \textcolor{purple}{
        Consequently, the \textbf{MSEF} block acts as a multilayer perceptron capable of performing efficient feature extraction on self-attended and denoised chrominance features, enabling high-quality restoration with minor parameter count increase.

    \begin{table*}
\centering
\small
\renewcommand{\arraystretch}{1.2}
\setlength{\tabcolsep}{7pt}
\begin{tabular}{l|cc|cc|cc|cc|cc}

\Xhline{1.0pt} % Thick horizontal line
\noalign{\vskip 1pt}
\Xhline{1pt}
\noalign{\vskip 1pt}

\rowcolor[gray]{0.92}
 & \multicolumn{2}{c|}{\textbf{Complexity}} & \multicolumn{2}{c|}{\textbf{LOL-v1}} & \multicolumn{2}{c|}{\textbf{LOL-v2-real}} & \multicolumn{2}{c|}{\textbf{LOL-v2-syn}} & \multicolumn{2}{c}{\textbf{SDSD}} \\ 
 
 \rowcolor[gray]{0.92}
                                 \multirow{-2}{*}{\textbf{Methods}} & \textbf{FLOPS (G)} & \textbf{Params (M)} & \textbf{PSNR} & \textbf{SSIM} & \textbf{PSNR} & \textbf{SSIM} & \textbf{PSNR} & \textbf{SSIM} & \textbf{PSNR} & \textbf{SSIM}        \\ 
\noalign{\vskip 1pt}
\Xhline{1.0pt}
% SID \cite{SID}                          & 13.73              & 7.76                & 14.35        & 0.436       & 13.24           & 0.442          & 15.04          & 0.610     & 24.09 & 0.698       \\ 
\noalign{\vskip 1pt}
3DLUT \cite{3DLUT}                       & \color{red}{0.075}              & \color{blue}{0.59}                & 21.35        & 0.585       & 20.19           & 0.745          & 22.17          & 0.854     & 21.78 & 0.652    \\ 
% DeepUPE \cite{DeepUPE}                      & 21.10              & 1.02                & 14.38        & 0.446       & 13.27           & 0.452          & 15.08          & 0.623     & 21.82 & 0.680          \\ 
% DeepLPF \cite{DeepLPF}                      & 5.86               & 1.77                & 15.28        & 0.473       & 14.10           & 0.480          & 16.02          & 0.587     & 22.49 & 0.661     \\ 
UFormer \cite{UFormer}                     & 12.00              & 5.29                & 16.36        & 0.771       & 18.82           & 0.771          & 19.66          & 0.871      & 23.51 & 0.804         \\ 
RetinexNet \cite{RetinexNet}                  & 587.47             & 0.84                & 18.92        & 0.427       & 18.32           & 0.447          & 19.09          & 0.774       & 20.90 & 0.623      \\ 
Sparse \cite{Sparse}                      & 53.26              & 2.33                & 17.20        & 0.640       & 20.06           & 0.816          & 22.05          & 0.905         & 24.27 & 0.834   \\ 
EnGAN \cite{EnGAN}                        & 61.01              & 114.35              & 20.00        & 0.691       & 18.23           & 0.617          & 16.57          & 0.734         & 20.06 & 0.610   \\ 
%RUAS \cite{RUAS}                         & {0.83}               & {0.003}               & 18.65        & 0.518       & 19.06           & 0.510          & 16.58          & 0.719         \\ 
FIDE \cite{FIDE}                         & 28.51              & 8.62                & 18.27        & 0.665       & 16.85           & 0.678          & 15.20          & 0.612         & 22.31 & 0.644 \\ 

\hline
\noalign{\vskip 1pt}
\textbf{LYT-Net} (Ours)                     & \color{blue}{3.49}              & \color{red}{0.045}               & \color{black}{22.38}       & {0.826}        & \color{black}{21.83}           & {0.849}          & \color{black}{23.78}          & \color{black}{0.921}   &    28.42        &   0.877  \\ 
\noalign{\vskip 1pt}
\Xhline{1pt}
\noalign{\vskip 1pt}
\Xhline{1pt}
\noalign{\vskip 1pt}

\rowcolor{orange!5}
KinD \cite{KinD}                        & 34.99              & 8.02                & 20.86        & 0.790       & 14.74           & 0.641          & 13.29          & 0.578         & \cellcolor{white} 21.96 & \cellcolor{white} 0.663   \\ 
\rowcolor{orange!5}
Restormer \cite{Restormer}               & 144.25             & 26.13               & 26.68        & 0.853       & 26.12           & {0.853}          & 25.43          & 0.859         & \cellcolor{white} 25.23 & \cellcolor{white} 0.815         \\ 
\rowcolor{orange!5}
DepthLux \cite{depthlux}               & -             & 9.75               & 26.06       & 0.793       & 26.16           & {0.794}          & 28.69          & 0.920         & \cellcolor{white} - & \cellcolor{white} -         \\ 
\rowcolor{orange!5}
ExpoMamba \cite{adhikarla2024expomamba}   & -             & 41            & 25.77       & \color{red}{0.860}        & 28.04       & \color{red}{0.885}           & -          & -         & \cellcolor{white} - & \cellcolor{white} -        \\ 
\rowcolor{orange!5}
MIRNet \cite{MIRNet}   & 785             & 31.76             & 26.52        & \color{blue}{0.856}        & 27.17       & 0.865           & 25.96          & 0.898         & \cellcolor{white} 25.76 &\cellcolor{white} 0.851        \\ 
\rowcolor{orange!5}
SNR-Net \cite{SNR-Net}                   & 26.35              & 4.01                & 26.72        & 0.851       & 27.21           & {0.871}          & 27.79          &  \color{red}{0.941}         & \cellcolor{white} \color{blue}{29.05} & \cellcolor{white} \color{blue}{0.880}         \\ 
\rowcolor{orange!5}
KAN-$\mathcal{T}$ \cite{kant}                    & -              & 2.80                & {26.66}        & {0.854}      & \color{red}{28.45}           & \color{blue}{0.884}          & {28.77}          & 0.939       & \cellcolor{white} - & \cellcolor{white} -  \\
\rowcolor{orange!5}
Retinexformer \cite{Retinexformer}                    & 15.57              & 1.61                & \color{blue}{27.14}        & 0.850      & {27.69}           & 0.856          & \color{blue}{28.99}          & 0.939       & \cellcolor{white} \color{red}{29.81} & \cellcolor{white} \color{red}{0.887}  \\  

\rowcolor{orange!5}
\hline
\noalign{\vskip 1pt}

\textbf{LYT-Net} (Ours)                        & \color{blue}{3.49}              & \color{red}{0.045}               & \color{red}{27.23}       & {0.853}        & \color{blue}{27.80}           & {0.873}          & \color{red}{29.38}          & \color{blue}{0.940}   &   \cellcolor{white}  28.42        &   \cellcolor{white} 0.877  \\ 

\noalign{\vskip 1pt}
\Xhline{1.0pt}
\noalign{\vskip 1pt}
\Xhline{1pt}
\end{tabular}
\vspace{3pt}
\caption{Quantitative results on LOL datasets. Best results are in {\color{red}red}, second best are in {\color{blue}blue}. \colorbox{orange!3}{Highlighted cells} show results with GT-Mean gamma correction~\cite{KinD}, which is widely used on the LOL datasets..}
\label{table:SOTA}
\vspace{-5mm}
\end{table*}

 \subsection{Loss Function}

        \vspace{0.5mm}
        In our approach, a hybrid loss function plays a pivotal role in training our model effectively. The hybrid loss $\mathbf{L}$ is formulated as in Eq.~\eqref{formula:loss_hybrid}, where $\alpha_1$ to $\alpha_5$ are hyperparameters used to balance each constituent loss function.
        
        \vspace{-0.5mm}
        \begin{equation}
        \small
        \mathbf{L}  = \mathbf{L}_{\text{S}} + \alpha_1 \mathbf{L}_{\text{Perc}} + \alpha_2 \mathbf{L}_{\text{Hist}} + \alpha_3 \mathbf{L}_{\text{PSNR}} + \alpha_4 \mathbf{L}_{\text{Color}} + \alpha_5 \mathbf{L}_{\text{MS-SSIM}}
        \label{formula:loss_hybrid}
        \vspace{-0.5mm}
        \end{equation}

        % - purple
        \vspace{-0.5mm}
        %zoinks
      
        The hybrid loss in our model combines several components to enhance image quality and perception. Smooth L1 loss $\mathbf{L}_{\text{S}}$ handles outliers by applying a quadratic or linear penalty based on the difference between predicted and true values. Perceptual loss $\mathbf{L}_{\text{Perc}}$ maintains feature consistency by comparing VGG-extracted feature maps. Histogram loss $\mathbf{L}_{\text{Hist}}$ aligns pixel intensity distributions between predicted and true images. PSNR loss $\mathbf{L}_{\text{PSNR}}$ reduces noise by penalizing mean squared error, while Color loss $\mathbf{L}_{\text{Color}}$ ensures color fidelity by minimizing differences in channel mean values. Lastly, Multiscale SSIM loss $\mathbf{L}_{\text{MS-SSIM}}$ preserves structural integrity by evaluating similarity across multiple scales. Together, these losses form a comprehensive strategy addressing various aspects of image enhancement.

\section{Results and Discussion}

%\section{EXPERIMENTS AND RESULTS}
\label{sec:experiments}

    \begin{figure*}[!h]
        \centering
        \includegraphics[width=0.98\linewidth]{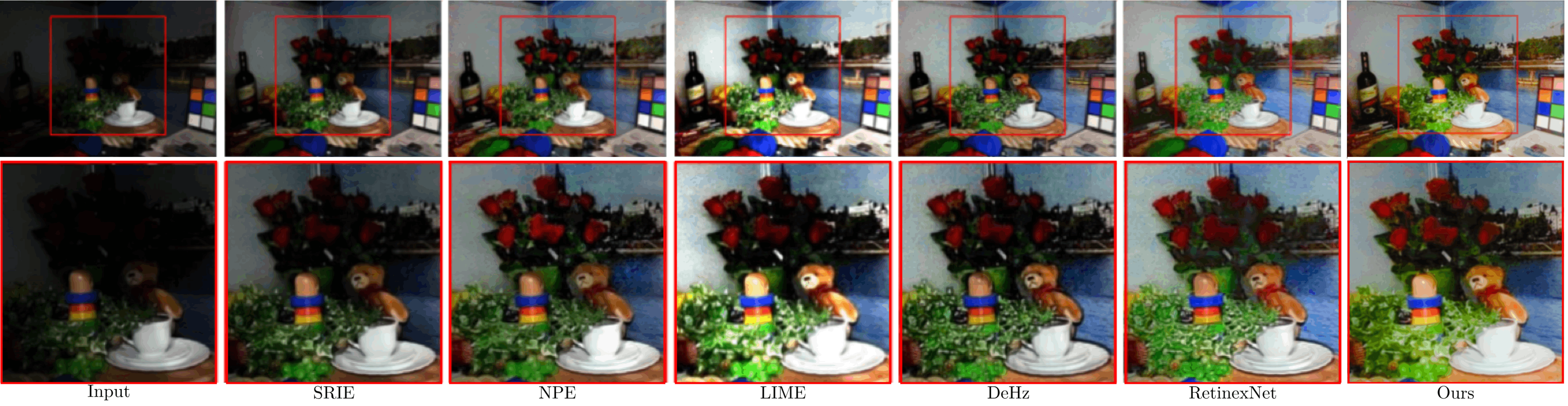}
        \vspace{-10pt}
        
        \caption{Qualitative comparison with SOTA LLIE methods on LIME dataset. Zoom-in regions are used to illustrate differences.}
        \label{fig:compare_zoom_2}
        \vspace{-4mm}
    \end{figure*}
    
% \subsection{Datasets usage}

%     \textbf{LYT-Net} is trained and evaluated on versions of the LOL dataset: v1, v2-real subset, and v2-synthetic subset. The training/testing splits corresponding to each version are 485:15 for LOL-v1, 689:100 for LOL-v2-real, and 900:100 for LOL-v2-synthetic.

%\subsection{Implementation specifics}

 \textbf{Implementation details:} The implementation of \textbf{LYT-Net} utilizes the TensorFlow framework. The ADAM Optimizer ($\beta_1 = 0.9$ and $\beta_2 = 0.999$) is employed for training over $1000$ epochs. The initial learning rate is set to $2 \times 10^{-4}$ and gradually decays to $1 \times 10^{-6}$ following a cosine annealing schedule, aiding in optimization convergence and avoiding local minima. The hyperparameters of the hybrid loss function are set as: $\alpha_1\!\!=\!\!0.06$, $\alpha_2\!\!=\!\!0.05$, $\alpha_3\!\!=\!\!0.0083$, $\alpha_4\!\!=\!\!0.25$, and $\alpha_5\!\!=\!\!0.5$.
    
    \vspace{-0.3mm}
    \textbf{LYT-Net} is trained and evaluated on: LOL-v1, LOL-v2-real, and LOL-v2-synthetic. The corresponding training/testing splits are $485:15$ for LOL-v1, $689:100$ for LOL-v2-real, and $900:100$ for LOL-v2-synthetic.
    
    \vspace{-0.5mm}
    During training, image pairs undergo random augmentations, including random cropping to $256 \times 256$ and random flipping/rotation, to prevent overfitting. The training is conducted with a batch size of $1$. Evaluation metrics include PSNR and SSIM for performance assessment.
    
    \vspace{1mm}

%\subsection{Quantitative results}
  
% \textcolor{blue}{
   \textbf{Quantitative results:} The proposed method is compared to SOTA LLIE techniques, as shown in Table~\ref{table:SOTA}, focusing on several aspects: quantitative performance on the LOL (LOL-v1, LOL-v2-real, LOL-v2-synthetic) and SDSD~\cite{sdsd} datasets, and model complexity.
% }

%zoinks

\vspace{-0.5mm}
   As shown in Table~\ref{table:SOTA}, \textbf{LYT-Net} consistently outperforms the current SOTA methods across all versions of the LOL dataset in terms of both PSNR and SSIM. Additionally, \textbf{LYT-Net} is highly efficient, requiring only 3.49G FLOPS and utilizing just 0.045M parameters, which gives it a significant advantage over other SOTA methods that are generally much more complex. The only exception is 3DLUT\cite{3DLUT}, which is comparable to our approach in terms of complexity. However, \textbf{LYT-Net} clearly surpasses the 3DLUT method in both PSNR and SSIM. This combination of strong performance and low complexity highlights the overall effectiveness of \textbf{LYT-Net}. 
   % \textcolor{purple}{
   On SDSD, where images are high resolution, our method shows limitations due to its significantly low parameter count. However, by utilizing a deeper variant of \textbf{LYT-Net}, we expect that performance increases accordingly.
   % }

    %\vspace{-0.5mm}
    %As indicated in Table~\ref{table:SOTA}, \textbf{LYT-Net} consistently ranks in the top three across all versions of the LOL dataset. In terms of complexity, \textbf{LYT-Net} is highly efficient, utilizing just $3.49G$ FLOPS and a minimal number of parameters ($0.045M$). This significantly reduces computational demands compared to other SOTA methods. The combination of strong performance and low complexity underlines the effectiveness of \textbf{LYT-Net}.
    
    %\vspace{-0.5mm}
    %Though \textbf{LYT-Net} ranks third in complexity, it outperforms models like 3DLUT and RUAS, which have lower computational requirements, by delivering superior quantitative results.
    
    %\vspace{-0.5mm}
    %Compared to models such as SNR-Net, Retinexformer, and MIRNet, \textbf{LYT-Net} achieves competitive performance while maintaining substantially lower computational costs.
    
    %\vspace{1mm}

%\subsection{Qualitative Results}

    %\vspace{-0.5mm}
    \textbf{Qualitative Results:} The qualitative evaluation of \textbf{LYT-Net} against SOTA LLIE techniques is shown in Fig.~\ref{fig:compare_LOL_v2_zoom} on the LOL dataset and in Fig.~\ref{fig:compare_zoom_2} on LIME~\cite{LIME}. 
    
    %zoinks
    
    \vspace{-0.5mm}
    Previous methods, such as KiND\cite{KinD}  and Restormer\cite{Restormer}, exhibit color distortion issues, as shown in Fig.~\ref{fig:compare_LOL_v2_zoom}. Additionally, several algorithms (e.g. MIRNet\cite{MIRNet}, and SNR-Net\cite{SNR-Net}) tend to produce over- or under-exposed areas, compromising image contrast while enhancing luminance. Similarly, Fig.~\ref{fig:compare_zoom_2} demonstrates that SRIE~\cite{SRIE}, DeHz~\cite{DeHz}, and NPE~\cite{NPE} result in a loss of contrast. In general, our \textbf{LYT-Net} is highly effective at enhancing visibility and low-contrast or poorly lit areas, while efficiently eliminating noise without introducing artifacts.

    %Previous methods, such as KiND and Restormer, exhibit color distortion issues, as seen in Fig.~\ref{fig:compare_LOL_v2_zoom}.
    
    % It's noteworthy that previous methods have displayed certain limitations. For instance, in Fig. \ref{fig:compare_MEF_v2_zoom}, LLNet\cite{LLNet}, AGLLNet \cite{AGLLNet} and RetinexNet exhibit issues with color distortion. Similarly, KiND and Restormer, as shown in Fig. \ref{fig:compare_LOL_v2_zoom}, also suffer from color distortion problems.
    
    %\vspace{-0.5mm}
    %Additionally, many algorithms tend to produce over- or under-exposed areas, which compromises image contrast while enhancing luminance. This issue is noticeable in algorithms like RUAS, MIRNet, and SNR-Net, as seen in Fig.~\ref{fig:compare_LOL_v2_zoom}. Similarly, Fig.~\ref{fig:compare_zoom_2} shows other algorithms such as SRIE~\cite{SRIE}, DeHz~\cite{DeHz}, and NPE~\cite{NPE}, where the enhancement leads to a loss of contrast.
    
    %\vspace{-0.5mm}
    
% \vspace{-3mm}

\vspace{-5pt}
\section{Ablation Study}
\vspace{-14pt}

%\section{EXPERIMENTS AND RESULTS}
\label{sec:ablation}

    .

The ablation study is conducted on the LOLv1 dataset, using PSNR 
{and CIEDE2000~\cite{ciede2000} as} 
quantitative metrics, and evaluates the impact of the \textbf{CWD} and \textbf{MSEF} blocks. In the YUV decomposition, applying \textbf{CWD} to the \textit{Y}-channel (used as the illumination map) results in the retention of lighting artifacts, leading to performance degradation compared to pooling operations and interpolation-based upsampling, which smoothen the illumination for better and more uniform lighting. However, \textbf{CWD} enhances the chrominance channels (\textit{U} and \textit{V}), preserving detail without introducing noise.
Moreover, the \textbf{MSEF} block consistently boosts performance across all \textbf{CWD} combinations, improving PSNR by 0.16, 0.24, and 0.26 dB, respectively, only increasing the parameter count by 546.

% \textcolor{purple}{
% The ablation study is conducted on the LOLv1 dataset, using PSNR {and CIEDE2000~\cite{ciede2000} as} the quantitative metrics, and investigates the impact of the \textbf{CWD} and \textbf{MSEF} blocks. Specifically, \textbf{CWD} is analyzed in the context of the YUV color space decomposition, where the \textit{Y}-channel serves as the illumination map. Although applying \textbf{CWD} to the \textit{Y}-channel effectively captures finer structures, it retains lighting artifacts, degrading performance compared to pooling and interpolation-based upsampling, which produce smoother and more uniform illumination. Nevertheless, \textbf{CWD} excels in enhancing the chrominance channels (\textit{U} and \textit{V}), preserving structural details without introducing additional noise.
% Furthermore, the \textbf{MSEF} block consistently boosts performance across all investigated \textbf{CWD} configurations, leading to PSNR improvements of 0.16, 0.24, and 0.26 dB, respectively, with a modest increase of only 546 parameters, making it a key component of our architectural design tailored for efficient illumination enhancement.
% }

\begin{table}[h]
\label{table:ablation}
\centering
\renewcommand{\arraystretch}{1.1} % Adjust vertical space between rows
\setlength{\tabcolsep}{4pt} % Adjust horizontal padding inside cells

% First table
\begin{minipage}{\linewidth}
    \centering
    \begin{tabular}{l|l|l|c|cc}
    \noalign{\vskip 1pt}
    \Xhline{1pt}
    \noalign{\vskip 1pt}
    \Xhline{1pt}
    \noalign{\vskip 1pt}
    \rowcolor[gray]{0.92}  % Apply light gray background to the header row
    \textbf{Y-CWD} & \textbf{UV-CWD} & \textbf{MSEF} & \textbf{Params} & \textbf{PSNR}$\uparrow$ & \textbf{CIEDE2000}$\downarrow$ \\
    \noalign{\vskip 1pt}
    \Xhline{1pt}
    \noalign{\vskip 1pt}
     \makecell{\checkmark} &  &  & 40238 & 26.62 & 6.3087 \\
     & \makecell{\checkmark} &  & 44377 & \color{black}{26.99} & 6.0148 \\
     \makecell{\checkmark} & \makecell{\checkmark} &  & 48516 & \color{black}{26.76} & 6.1975 \\ \hline
     \makecell{\checkmark} &  & \makecell{\checkmark} & 40784 & 26.78 & 6.1816 \\
     & \makecell{\checkmark} & \makecell{\checkmark} & 44923 & \color{red}{27.23} & 5.8242 \\
     \makecell{\checkmark} & \makecell{\checkmark} & \makecell{\checkmark} & 49062 & \color{blue}{27.02} & 5.9910 \\
    \noalign{\vskip 1pt}
     \Xhline{1pt}
    \noalign{\vskip 1pt}
    \Xhline{1pt}
    \end{tabular}
    \vspace{3pt}
    \parbox{\linewidth}{\centering \caption{Ablation study: Performance and parameter impact of CWD and MSEF blocks.}}
\end{minipage}

% \vspace{12pt} % Add some vertical space between the tables

% Second table
% \begin{minipage}{\linewidth}
%     \centering
%     \begin{tabular}{l|c|c}
%     \Xhline{1pt}
%     \textbf{Methods} & \textbf{Params} & \textbf{PSNR} \\ \hline
%     w/ MSEF & 44923 & \textbf{27.23} \\
%     w/o MSEF & 44377 & 26.76 \\
%     \Xhline{1pt}
%     \end{tabular}
%     \vspace{3pt}
%     \parbox{\linewidth}{\centering \caption{Performance and parameter impact of MSEF applied on chrominance (U+V)}}
% \end{minipage}
\vspace{-14mm}
\end{table}

% The ablation study is performed on LOLv1 and utilizes PSNR as quantitative measurement and analyzes the impact of the \textbf{CWD} and \textbf{MSEF} blocks. In the YUV decomposition, applying \textbf{CWD} to the \textit{Y}-channel (used as the illumination map) determines retention of lighting artifacts, leading to performance drops compared to pooling operations and interpolation-based upsampling, which smoothen the illumination for better results. Conversely, \textbf{CWD} benefits the chrominance (\textit{U} and \textit{V}), retaining detail without introducing noise.  The \textbf{MSEF} block consistently improves performance across all \textbf{CWD} combinations by 1.29, 1.09 and 1.05 dB PSNR respectively, while increasing parameter count by only around 600, making it an efficient enhancement.

\vspace{-5pt}
\section{Conclusions}
\vspace{3pt}

%\section{CONCLUSION AND FUTURE WORKS}
\label{sec:conclusion}
%zoinks
\vspace{-4pt}
We introduce \textbf{LYT-Net}, an innovative lightweight transformer-based model for enhancing low-light images. Our approach utilizes a dual-path framework, processing chrominance and luminance separately to improve the model's ability to manage illumination adjustments and restore corrupted regions. \textbf{LYT-Net} integrates multiple layers and modular blocks, including two unique \textbf{CWD} and \textbf{MSEF} — as well as the traditional ViT block with \textbf{MHSA}. A comprehensive qualitative and  quantitative analysis demonstrates that \textbf{LYT-Net} consistently outperforms SOTA methods on all versions of the LOL dataset in terms of PSNR and SSIM, while maintaining high computational efficiency.

\noindent{\textbf{Acknowledgement}}: Part of this research is supported by the "Romanian Hub for Artificial Intelligence -- HRIA", Smart Growth, Digitization and Financial Instruments Program, 2021-2027, MySMIS no. 334906.

    %The data presented in these work highlights \textbf{LYT-Net}'s ability to deliver high-quality image enhancement with minimal computational resources. Its performance is comparable to, and in some cases surpasses, heavier and more complex models, making it an attractive solution for applications where computational efficiency is as crucial.
    %In summary, \textbf{LYT-Net} distinguishes itself by offering a highly effective balance between performance and efficiency, achieving top-tier results on standard benchmarks while maintaining ultra-lightweight properties.
    %Looking ahead, we plan to further evaluate our model on larger datasets and incorporate user feedback to enhance our benchmarking. Given \textbf{LYT-Net}'s low complexity, we also foresee its potential integration with sensor systems, expanding its applicability in real-world scenarios.

% For peer review papers, you can put extra information on the cover
% page as needed:
% \ifCLASSOPTIONpeerreview
% \begin{center} \bfseries EDICS Category: 3-BBND \end{center}
% \fi
%
% For peerreview papers, this IEEEtran command inserts a page break and
% creates the second title. It will be ignored for other modes.
\IEEEpeerreviewmaketitle

% if have a single appendix:
%\appendix[Proof of the Zonklar Equations]
% or
%\appendix  % for no appendix heading
% do not use \section anymore after \appendix, only \section*
% is possibly needed

% use appendices with more than one appendix
% then use \section to start each appendix
% you must declare a \section before using any
% \subsection or using \label (\appendices by itself
% starts a section numbered zero.)
%

% use section* for acknowledgment
%\section*{Acknowledgment}

%The authors would like to thank...

% Can use something like this to put references on a page
% by themselves when using endfloat and the captionsoff option.
\ifCLASSOPTIONcaptionsoff
  \newpage
\fi

% trigger a \newpage just before the given reference
% number - used to balance the columns on the last page
% adjust value as needed - may need to be readjusted if
% the document is modified later
%\IEEEtriggeratref{8}
% The "triggered" command can be changed if desired:
%\IEEEtriggercmd{\enlargethispage{-5in}}

% references section

\bibliographystyle{IEEEtran}
\bibliography{ref.bib}

% that's all folks
\end{document}